\documentclass[11pt,a4paper]{article}
\usepackage[hyperref]{acl2019}
\usepackage{times}
\usepackage{latexsym}

\usepackage{url}

\aclfinalcopy %

\usepackage[utf8]{inputenc}
\usepackage{graphicx}
\usepackage{multirow}
\usepackage{booktabs}
\usepackage{amsmath}
\DeclareMathOperator{\score}{score}
\DeclareMathOperator{\nn}{NN}
\DeclareMathOperator{\margin}{margin}

\title{Margin-based Parallel Corpus Mining \\ with Multilingual Sentence Embeddings}

\author{Mikel Artetxe \\
  University of the Basque Country (UPV/EHU)\thanks{This work was performed during an internship at Facebook AI Research.} \\
  {\tt mikel.artetxe@ehu.eus} \\\And
  Holger Schwenk \\
  Facebook AI Research \\
  {\tt schwenk@fb.com} \\}

\date{}

\begin{document}
\maketitle
\begin{abstract}
Machine translation is highly sensitive to the size and quality of the training data, which has led to an increasing interest in collecting and filtering large parallel corpora. In this paper, we propose a new method for this task based on multilingual sentence embeddings.
In contrast to previous approaches, which rely on nearest neighbor retrieval with a hard threshold over cosine similarity, our proposed method accounts for the scale inconsistencies of this measure, considering the margin between a given sentence pair and its closest candidates instead. Our experiments show large improvements over existing methods. We outperform the best published results on the BUCC mining task and the UN reconstruction task by more than 10 F1 and 30 precision points, respectively. Filtering the English-German ParaCrawl corpus with our approach, we obtain 31.2 BLEU points on newstest2014, an improvement of more than one point over the best official filtered version.
\end{abstract}

\newcommand{\InsertExample}{
\begin{table*}[t]
\begin{small}
\begin{center}
  \begin{tabular}{cl}
    \toprule
    (A) & \textit{Les produits agricoles sont constitués de thé, de riz, de sucre, de tabac, de camphre, de fruits et de soie.} \\
    \midrule
    0.818 & Main crops include wheat, sugar beets, potatoes, cotton, tobacco, vegetables, and fruit. \\
    0.817 & The fertile soil supports wheat, corn, barley, tobacco, sugar beet, and soybeans. \\
    0.814 & Main agricultural products include grains, cotton, oil, pigs, poultry, fruits, vegetables, and edible fungus. \\
    0.808 & The important crops grown are cotton, jowar, groundnut, rice, sunflower and cereals. \\
    \bottomrule
    \\
    \toprule
    (B) & \textit{Mais dans le contexte actuel, nous pourrons les ignorer sans risque.} \\
    \midrule
    0.737 & But, in view of the current situation, we can safely ignore these. \\
    0.499 & But without the living language, it risks becoming an empty shell. \\
    0.498 & While the risk to those working in ceramics is now much reduced, it can still not be ignored. \\
    0.488 & But now they have discovered they are not free to speak their minds. \\
    \bottomrule
  \end{tabular}
\end{center}
\end{small}
\caption{Motivating example of the proposed method. We show the nearest neighbors of two French sentences on the BUCC training set along with their cosine similarities. Only the nearest neighbor of B is a correct translation, yet that of A has a higher cosine similarity. We argue that this is caused by the cosine similarity of different sentences being in different scales, making it a poor indicator of the confidence of the prediction. Our method tackles this issue by considering the margin between a given candidate and the rest of the $k$ nearest neighbors.}
\label{tab:motivation}
\end{table*}
}

\newcommand{\InsertFigArchi}{
\begin{figure*}[t] \centering
\includegraphics[width=0.95\textwidth]{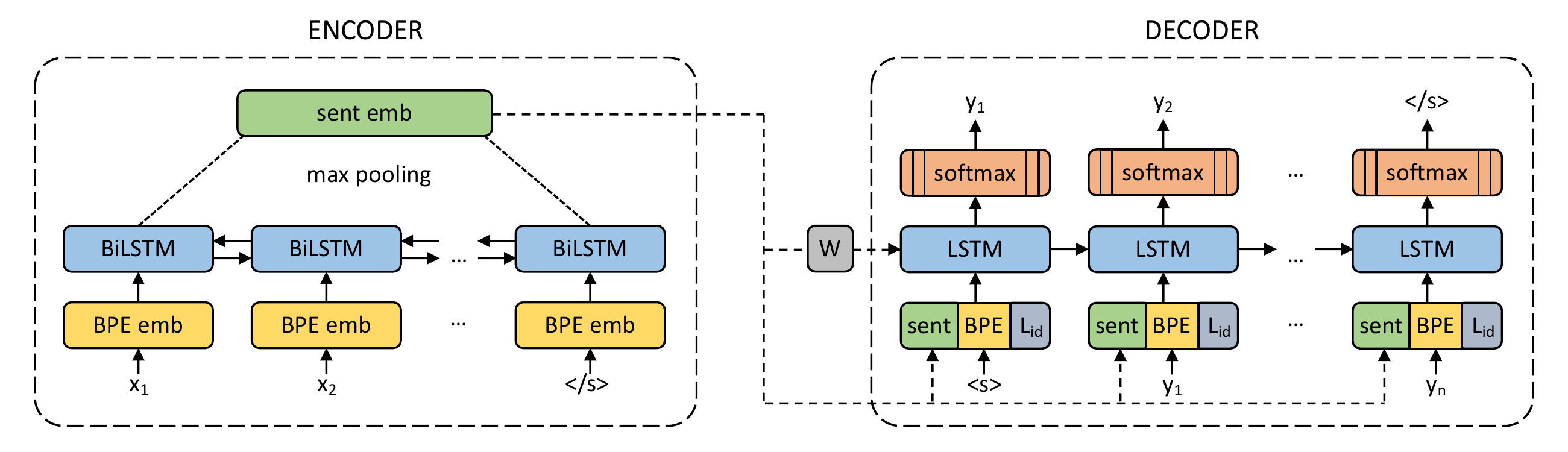}
\caption{Architecture of our system to learn multilingual sentence embeddings.}
\label{fig:architecture}
\end{figure*}
}

\section{Introduction}
\label{sec:introduction}

While Neural Machine Translation (NTM) has obtained breakthrough improvements in standard benchmarks, it is known to be particularly sensitive to the size and quality of the training data \citep{koehn2017six,khayrallah2018impact}. In this context, effective approaches to mine and filter parallel corpora are crucial to apply NMT in practical settings.

Traditional parallel corpus mining has relied on heavily engineered systems. Early approaches were mostly based on metadata information from web crawls \citep{resnik1999mining,shi2006dom}. More recent methods focus on the textual content instead. For instance, Zipporah learns a %
classifier over bag-of-word features to distinguish between ground truth translations and synthetic noisy ones \citep{xu2017zipporah}. STACC uses seed \mbox{lexical} translations induced from IBM alignments, which are combined with set expansion operations to score translation candidates through the Jaccard similarity coefficient \citep{etchegoyhen2016set,azpeitia2017weighted,azpeitia2018extracting}. Many of these approaches rely on cross-lingual document retrieval \citep{utiyama2003reliable,munteanu2005improving,munteanu2006extracting,rauf2009comparable} or machine translation \citep{rauf2009comparable,bouamor2018h2}.%

More recently, a new research line has shown promising results using multilingual sentence embeddings alone\footnote{Multilingual entence embeddings have also been used as part of a larger system, either to obtain an initial alignment that is then further filtered \citep{bouamor2018h2} or as an intermediate representation of an end-to-end classifier \citep{gregoire2017bucc}.} \citep{schwenk2018filtering,guo:2018:wmt_effective}.
These methods use an NMT inspired encoder-decoder to train sentence embeddings on existing parallel data, which are then directly applied to retrieve and filter new parallel sentences using nearest neighbor retrieval over cosine similarity with a hard threshold \citep{espana2017empirical,hassan2018achieving,schwenk2018filtering}.

\InsertExample

In this paper, we argue that this retrieval method suffers from the scale of cosine similarity not being globally consistent. As illustrated by the example in Table \ref{tab:motivation}, some sentences without any correct translation have overall high cosine scores, making them rank higher than other sentences with a correct translation. This issue was also pointed out by \citet{guo:2018:wmt_effective}, who learn an encoder to score known translation pairs above synthetic negative examples and train a separate model to dynamically scale and shift the dot product on held out supervised data. In contrast, our proposed method tackles this issue by considering the margin between the cosine of a given sentence pair and that of its respective $k$ nearest neighbors.

\section{Multilingual sentence embeddings}
\label{sec:embeddings}

Figure \ref{fig:architecture} shows our encoder-decoder architecture to learn multilingual sentence embeddings, which is based on \citet{schwenk2018filtering}. The encoder consists of a bidirectional LSTM, and our sentence embeddings are obtained by applying a max-pooling operation over its output. These embeddings are fed into an LSTM decoder in two ways: 1)~they are used to initialize its hidden and cell state after a linear transformation, and 2) they are concatenated to the input embeddings at every time step. We use a shared encoder and decoder for all languages with a joint 40k BPE vocabulary learned on the concatenation of all training corpora.\footnote{Prior to BPE segmentation, we tokenize and lowercase the input text using standard Moses tools. As the only exception, we use Jieba (\url{https://github.com/fxsjy/jieba}) for Chinese word segmentation.} The encoder is fully language agnostic, without any explicit signal of the input or output language, whereas the decoder receives an output language ID embedding at every time step. Training minimizes the cross-entropy loss on parallel corpora, alternating over all combinations of the languages involved. We train on 4 GPUs with a total batch size of 48,000 tokens, using Adam with a learning rate of 0.001 and dropout set to 0.1. We use a single layer for both the encoder and the decoder with a hidden size of 512 and 2048, respectively, yielding 1024 dimensional sentence embeddings. The input embeddings size is set to 512, while the language ID embeddings have 32 dimensions. After training, the decoder is discarded, and the encoder is used to map a sentence to a fixed-length vector.

\vspace{-0pt}
\section{Scoring and filtering parallel sentences}
\label{sec:mining}

The multilingual encoder can be used to mine parallel sentences by taking the nearest neighbor of each source sentence in the target side according to cosine similarity, and filtering those below a fixed threshold. While this approach has been reported to be competitive \citep{schwenk2018filtering}, we argue that it suffers from the scale of cosine similarity not being globally consistent across different sentences.\footnote{Note that, even if cosine similarity is normalized in the (-1, 1) range,
it is still susceptible to concentrate around different values.} For instance, Table~\ref{tab:motivation} shows an example where an incorrectly aligned sentence pair has a larger cosine similarity than a correctly aligned one, thus making it impossible to filter it through a fixed threshold. In that case, all four nearest neighbors have equally high values. In contrast, for example B, there is a big gap between the nearest neighbor and its other candidates. As such, we argue that the margin between the similarity of a given candidate and that of its $k$ nearest neighbors is a better indicator of the strength of the alignment.\footnote{As a downside, this approach will penalize sentences with many paraphrases in the corpus. While possible, we argue that such cases rarely happen in practice and, even when they do, filtering them is unlikely to cause any major harm.} We next describe our scoring method inspired by this idea in Section~\ref{subsec:scoring}, and discuss our candidate generation and filtering strategy in Section~\ref{subsec:filtering}.

\InsertFigArchi

\subsection{Margin-based scoring} \label{subsec:scoring}

We consider the margin between the cosine of a given candidate and the average cosine of its $k$ nearest neighbors in both directions as follows:
\begin{multline*}
    \score(x, y) = \margin (\cos(x, y), \\
    \sum_{z \in \nn_k(x)}{\frac{\cos(x, z)}{2k}} +  \sum_{z \in \nn_k(y)}{\frac{\cos(y, z)}{2k}})
\end{multline*}
where $\nn_k(x)$ denotes the $k$ nearest neighbors of $x$ in the other language excluding duplicates,\footnote{Unless otherwise indicated, we use $k=4$.} and analogously for $\nn_k(y)$. We explore the following variants of this general definition:

\begin{itemize}
    \item \textbf{Absolute} ($\margin(a, b) = a$): Ignoring the average. This is equivalent to cosine similarity and thus our baseline.
    \item \textbf{Distance} ($\margin(a, b) = a - b$): Subtracting the average cosine similarity from that of the given candidate. This is proportional to the CSLS score \citep{conneau2018word}, which was originally motivated to mitigate the hubness problem on Bilingual Lexicon Induction (BLI) over cross-lingual word embeddings.\footnote{While our work is motivated by thresholding, which is not used in BLI, this connection points out a related problem that our approach also addresses: even when the source sentence is fixed, the potentially different scales of its target candidates might also affect their relative ranking, which ultimately causes the hubness problem. Thanks to its bidirectional nature, our proposed scoring method penalizes target sentences with overall high cosine similarities, so it can learn better alignments that account for this factor.}
    \item \textbf{Ratio} ($\margin(a, b) = \frac{a}{b}$): The ratio between the candidate and the average cosine of its nearest neighbors in both directions.
\end{itemize}

\subsection{Candidate generation and filtering} \label{subsec:filtering}

When mining parallel sentences, we explore the following strategies to generate candidates:
\begin{itemize}
    \item \textbf{Forward}: Each source sentence is aligned with exactly one best scoring target sentence.\footnote{For efficiency, only the $k$ nearest neighbors over cosine similarity are considered, where the neighborhood size $k$ is the same as that used for the margin-based scoring.} Some target sentences may be aligned with multiple source sentences or with none.
    \item \textbf{Backward}: Equivalent to the forward strategy, but going in the opposite direction.
    \item \textbf{Intersection} of forward and backward candidates, which discards sentences with inconsistent alignments.
    \item \textbf{Max. score}: Combination of forward and backward candidates that, instead of discarding all inconsistent alignments, it selects those with the highest score.
\end{itemize}

These candidates are then sorted according to their margin scores, and a threshold is applied. This can be either optimized on the development data, or adjusted to obtain the desired corpus size.

\section{Experiments and results}
\label{sec:experiments}

We next present our results on the BUCC mining task, UN corpus reconstruction, and machine translation over filtered ParaCrawl. All experiments use an English/French/Spanish/German multilingual encoder trained on Europarl v7 \cite{Koehn:2005:mtsummit_eurparl} for 10 epochs. To cover all languages in BUCC, we use a separate English/French/Russian/Chinese model trained on the UN corpus \citep{ziemski2016united} for 4 epochs.

\begin{table}[t]
\begin{small}
\begin{center}
  \addtolength{\tabcolsep}{-3pt}
  \begin{tabular}{clrrrrrrr}
    \toprule
    \multirow{2}{*}{\bf Func.} & \multirow{2}{*}{\bf Retrieval} & \multicolumn{3}{c}{\bf EN-DE} & & \multicolumn{3}{c}{\bf EN-FR} \\
    \cmidrule{3-5} \cmidrule{7-9}
    & & \multicolumn{1}{c}{\bf P} & \multicolumn{1}{c}{\bf R} & \multicolumn{1}{c}{\bf F1} & & \multicolumn{1}{c}{\bf P} & \multicolumn{1}{c}{\bf R} & \multicolumn{1}{c}{\bf F1} \\
    \midrule
    \multirow{4}{*}{\shortstack{Abs. \\ (cos)}}
    & Forward & 78.9 & 75.1 & 77.0 & & 82.1 & 74.2 & 77.9 \\
    & Backward & 79.0 & 73.1 & 75.9 & & 77.2 & 72.2 & 74.7 \\
    & Intersection & 84.9 & 80.8 & 82.8 & & 83.6 & 78.3 & 80.9 \\
    & Max. score & 83.1 & 77.2 & 80.1 & & 80.9 & 77.5 & 79.2 \\
    \midrule
    \multirow{4}{*}{Dist.}
    & Forward & 94.8 & 94.1 & 94.4 & & 91.1 & \bf 91.8 & 91.4 \\
    & Backward & 94.8 & 94.1 & 94.4 & & 91.5 & 91.4 & 91.4 \\
    & Intersection & 94.9 & 94.1 & 94.5 & & 91.2 & \bf 91.8 & 91.5 \\
    & Max. score & 94.9 & 94.1 & 94.5 & & 91.2 & \bf 91.8 & 91.5 \\
    \midrule
    \multirow{4}{*}{\shortstack{Ratio}}
    & Forward & 95.2 & \bf 94.4 & \bf 94.8 & & \bf 92.4 & 91.3 & 91.8 \\
    & Backward & 95.2 & \bf 94.4 & \bf 94.8 & & 92.3 & 91.3 & 91.8 \\
    & Intersection & \bf 95.3 & \bf 94.4 & \bf 94.8 & & \bf 92.4 & 91.3 & \bf 91.9 \\
    & Max. score & \bf 95.3 & \bf 94.4 & \bf 94.8 & & \bf 92.4 & 91.3 & \bf 91.9 \\
    \bottomrule
  \end{tabular}
\end{center}
\end{small}
\caption{BUCC results (precision, recall and F1) on the training set, used to optimize the filtering threshold.}
\label{tab:results_ablation}
\end{table}

\subsection{BUCC mining task} \label{subsec:bucc}

The shared task of the workshop on Building and Using Comparable Corpora (BUCC) is a well-established evaluation framework for bitext mining \citep{zweigenbaum2017overview,zweigenbaum2018overview}. The task is to mine for parallel sentences between English and four foreign languages: German, French, Russian and Chinese. There are 150K to 1.2M sentences for each language, split into a sample, training and test set. About 2--3\% of the sentences are parallel.

Table \ref{tab:results_ablation} reports precision, recall and F1 scores on the training set.\footnote{Note that the gold standard information was exclusively used to optimize the filtering threshold for each configuration, making results comparable across different variants.} Our results show that multilingual sentence embeddings already achieve competitive performance using standard forward retrieval over cosine similarity, which is in line with \citet{schwenk2018filtering}. Both of our bidirectional retrieval strategies achieve substantial improvements over this baseline while still relying on cosine similarity, with \textit{intersection} giving the best results.
Moreover, our proposed margin-based scoring brings large improvements when using either the \textit{distance} or the \textit{ratio} functions, outperforming cosine similarity by more than 10 points in all cases. The best results are achieved by \textit{ratio}, which outperforms \textit{distance} by 0.3-0.5 points. Interestingly, the retrieval strategy has a very small effect in both cases, suggesting that the proposed scoring is more robust than cosine.%

\begin{table}[t]
\begin{small}
\begin{center}
  \addtolength{\tabcolsep}{-3pt}
  \begin{tabular}{lcccc}
    \toprule
    & \multicolumn{1}{c}{\bf en-de} & \multicolumn{1}{c}{\bf en-fr} & \multicolumn{1}{c}{\bf en-ru} & \multicolumn{1}{c}{\bf en-zh} \\
    \midrule
    \citet{azpeitia2017weighted} & 83.7 & 79.5 & - & - \\
    \citet{azpeitia2018extracting} & 85.5 & 81.5 & 81.3 & 77.5 \\
    \citet{bouamor2018h2} & - & 76.0 & - & - \\
    \citet{schwenk2018filtering} & 76.9 & 75.8 & 73.8 & 71.6 \\
    \midrule
    Proposed method (Europarl) & \bf 95.6 & \bf 92.9 & - & - \\
    Proposed method (UN) & - & - & \bf 92.0 & \bf 92.6 \\
    \bottomrule
  \end{tabular}
\end{center}
\end{small}
\caption{BUCC results (F1) on the test set. We use the \textit{ratio} function with \textit{maximum score} retrieval and the filtering threshold optimized on the training set.}
\label{tab:results_bucc}
\end{table}

Table \ref{tab:results_bucc} reports the results on the test set for both the Europarl and the UN model in comparison to previous work.\footnote{We use the \textit{ratio} margin function with \textit{maximum score} retrieval for our method. The filtering threshold was optimized to maximize the F1 score on the training set for each language pair and model. The gold-alignments of the test set are not publicly available -- these scores on the test set are calculated by the organizers of the BUCC workshop. We have done one single submission.} Our proposed system outperforms all previous methods by a large margin, obtaining improvements of 10-15 F1 points and showing very consistent performance across different languages, including distant ones.

\subsection{UN corpus reconstruction} \label{subsec:un}

So as to compare our method to the similarly motivated system of \citet{guo:2018:wmt_effective}, we mimic their experiment on aligning the 11.3M sentences of the UN corpus. This task does not require any filtering, so we use \textit{forward} retrieval with the \textit{ratio} margin function. As shown in Table \ref{tab:results_un}, our system outperforms that of \citet{guo:2018:wmt_effective} by a large margin despite using only a fraction of the training data (2M sentences from Europarl in contrast with over 400M sentences from Google's internal data).

\begin{table}[t]
\begin{small}
\begin{center}
  \begin{tabular}{lcc}
    \toprule
    & \bf en-fr & \bf en-es \\
    \midrule
    \citet{guo:2018:wmt_effective} & 48.90 & 54.94 \\
    \midrule
    Proposed method & \bf 83.27 & \bf 85.78 \\
    \bottomrule
  \end{tabular}
\end{center}
\end{small}
\caption{Results on UN corpus reconstruction (P@1)}
\label{tab:results_un}
\end{table}

\subsection{Filtering ParaCrawl for NMT} \label{subsec:paracrawl}

Finally, we filter the English-German ParaCrawl corpus and evaluate NMT models trained on them. Our NMT models use \texttt{fairseq}'s implementation of the big transformer model \citep{vaswani2017attention}, using the same configuration as \citet{Ott:2018:wmt_scale_nmt} and training for 100 epochs. Following common practice, we use newstest2013 and newstest2014 as our development and test sets, respectively, and report both tokenized and detokenized BLEU scores as computed by \texttt{multi-bleu.perl} and sacreBLEU. %
We decode with a beam size of 5 using an ensemble of the last 10 epochs. One single model is only slightly worse.

\begin{figure}[t] \centering
\includegraphics[width=0.40\textwidth]{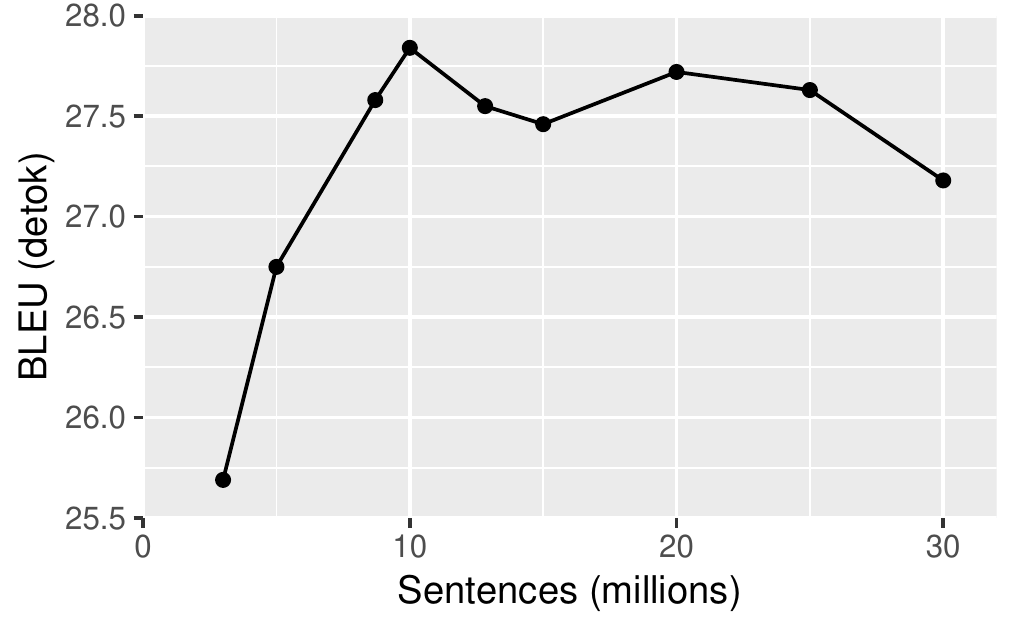}
\caption{English-German Dev results (newstest2013) using different thresholds to filter ParaCrawl.}
\label{fig:mt_dev}
\end{figure}

Given the large size of ParaCrawl, we first preprocess it to remove all duplicated sentence pairs, sentences for which the fastText language identification model\footnote{\url{https://fasttext.cc/docs/en/language-identification.html}} predicts a different language, those with less than 3 or more than 80 tokens, or those with either an overlap of at least 50\% or a ratio above 2 between the source and target tokens. This reduces the corpus size from 4.59 billion to 64.4 million sentence pairs, mostly due to deduplication. We then score each sentence pair with the \textit{ratio} function, processing the entire corpus in batches of 5 million sentences, and take the top scoring entries up to the desired size. Figure \ref{fig:mt_dev} shows the development BLEU scores of the resulting system for different thresholds, which peaks at 10 million sentences. As shown in Table~\ref{tab:results_paracrawl}, this model clearly outperforms the two official filtered versions of ParaCrawl in the test set.

\begin{table}[t]
\begin{small}
\begin{center}
  \begin{tabular}{lcccc}
    \toprule
    & \multirow{2}{*}{\bf \#SENT} & & \multicolumn{2}{c}{\bf BLEU} \\
    \cmidrule{4-5}
    & & & \bf tok & \bf detok \\
    \midrule
    BiCleaner v1.2 & 17.4M & & 30.05 & 29.37 \\
    Zipporah v1.2 & 40.5M & & 24.78 & 24.38 \\
    \midrule
    Proposed method & 10.0M & & \bf 31.19 & \bf 30.53 \\
    \bottomrule
  \end{tabular}
\end{center}
\end{small}
\caption{Results on English-German newstest2014 for different filtered versions of the ParaCrawl corpus.}
\label{tab:results_paracrawl}
\end{table}

Finally, Table \ref{tab:results_mt_sota} compares our results to previous works in the literature using different training data. In addition to our ParaCrawl system, we include an additional one combining it with all parallel data from WMT18 except CommonCrawl. As it can be seen, our system outperforms all previous systems but \citet{edunov:2018:emnlp_backtrans}, who use a large in-domain monolingual corpus through back-translation, making both works complementary. Quite remarkably, our full system outperforms \citet{Ott:2018:wmt_scale_nmt} by nearly 2 points despite using the same configuration and training data, so our improvement can be attributed to a better filtering of ParaCrawl.\footnote{To confirm this, we trained a separate model on WMT data, obtaining 29.4 tokenized BLEU. This is on par with the results reported by \citet{Ott:2018:wmt_scale_nmt} for the same data (29.3 tokenized BLEU). This shows that the difference cannot be attributed to implementation details.}

\begin{table}[t]
\begin{small}
\begin{center}
  \begin{tabular}{lccclll}
    \toprule
    & \multirow{2}{*}{\bf DATA} & \multicolumn{2}{c}{\bf BLEU} \\
    \cmidrule{3-4}
    & & \multicolumn{1}{c}{\bf tok} & \multicolumn{1}{c}{\bf detok} \\
    \midrule
    \citet{wu2016google} & wmt & 26.3 & - \\
    \citet{Gehring:2017:fairseq_icml} & wmt & 26.4 & - \\
    \citet{vaswani2017attention} & wmt & 28.4 & - \\
    \citet{ahmed2017weighted} & wmt & 28.9 & - \\
    \citet{shaw:2018:naacl_selfattn} & wmt & 29.2 & - \\
    \citet{Ott:2018:wmt_scale_nmt} & wmt & 29.3 & 28.6 \\
    \citet{Ott:2018:wmt_scale_nmt} & wmt+pc & 29.8 & 29.3 \\
    \citet{edunov:2018:emnlp_backtrans} & wmt+nc & 35.0 & 33.8 \\
    \midrule
    \multirow{2}{*}{Proposed method} &  pc & 31.2 & 30.5 \\
    & wmt+pc & 31.8 & 31.1 \\
    \bottomrule
  \end{tabular}
\end{center}
\end{small}
\caption{Results on English-German newstest2014 in comparison to previous work. \textit{wmt} for WMT parallel data (excluding ParaCrawl), \textit{pc} for ParaCrawl, and \textit{nc} for monolingual News Crawl with back-translation.}
\label{tab:results_mt_sota}
\end{table}

\section{Conclusions and future work}
\label{sec:conclusions}

In this paper, we propose a new method for parallel corpus mining based on multilingual sentence embeddings.
We use a sequence-to-sequence architecture to train a multilingual sentence encoder on an initial parallel corpus, and a novel margin-based scoring method that overcomes the scale inconsistencies of cosine similarity.

Our experiments show large improvements over previous methods. Our system obtains the best published results on the BUCC mining task, outperforming previous systems by more than 10 F1 points for all the four language pairs. In addition, our method obtains up to 85\% precision at reconstructing the 11.3M sentence pairs from the UN corpus, improving over the similarly motivated method of \citet{guo:2018:wmt_effective} by more than 30 points. Finally, we show that our improvements also carry over to downstream machine translation, as we obtain 31.2 BLEU points for English-German newstest2014 training on our filtered version of ParaCrawl, an improvement of more than one point over the best performing official release.

The code of this work is freely available as part of the LASER toolkit, together with an additional single encoder which covers 93 languages.\footnote{\url{https://github.com/facebookresearch/LASER}}

\bibliography{acl2019}

\begin{thebibliography}{30}
\expandafter\ifx\csname natexlab\endcsname\relax\def\natexlab#1{#1}\fi

\bibitem[{Abdul-Rauf and Schwenk(2009)}]{rauf2009comparable}
Sadaf Abdul-Rauf and Holger Schwenk. 2009.
\newblock \href {http://www.aclweb.org/anthology/E09-1003} {{On the Use of
  Comparable Corpora to Improve SMT performance}}.
\newblock In \emph{EACL}, pages 16--23.

\bibitem[{Ahmed et~al.(2017)Ahmed, Keskar, and Socher}]{ahmed2017weighted}
Karim Ahmed, Nitish~Shirish Keskar, and Richard Socher. 2017.
\newblock {Weighted Transformer Network for Machine Translation}.
\newblock \emph{arXiv:1711.02132}.

\bibitem[{Azpeitia et~al.(2017)Azpeitia, Etchegoyhen, and
  Mart{\'i}nez~Garcia}]{azpeitia2017weighted}
Andoni Azpeitia, Thierry Etchegoyhen, and Eva Mart{\'i}nez~Garcia. 2017.
\newblock \href {http://aclweb.org/anthology/W17-2508} {{Weighted Set-Theoretic
  Alignment of Comparable Sentences}}.
\newblock In \emph{BUCC}, pages 41--45.

\bibitem[{Azpeitia et~al.(2018)Azpeitia, Etchegoyhen, and
  Mart{\'i}nez~Garcia}]{azpeitia2018extracting}
Andoni Azpeitia, Thierry Etchegoyhen, and Eva Mart{\'i}nez~Garcia. 2018.
\newblock {Extracting Parallel Sentences from Comparable Corpora with STACC
  Variants}.
\newblock In \emph{BUCC}.

\bibitem[{Bouamor and Sajjad(2018)}]{bouamor2018h2}
Houda Bouamor and Hassan Sajjad. 2018.
\newblock {H2@BUCC18: Parallel Sentence Extraction from Comparable Corpora
  Using Multilingual Sentence Embeddings}.
\newblock In \emph{BUCC}.

\bibitem[{Conneau et~al.(2018)Conneau, Lample, Ranzato, Denoyer, and
  J{\'{e}}gou}]{conneau2018word}
Alexis Conneau, Guillaume Lample, Marc'Aurelio Ranzato, Ludovic Denoyer, and
  Herv{\'{e}} J{\'{e}}gou. 2018.
\newblock \href {https://openreview.net/pdf?id=H196sainb} {{Word Translation
  Without Parallel Data}}.
\newblock In \emph{ICLR}.

\bibitem[{Edunov et~al.(2018)Edunov, Ott, Auli, and
  Grangier}]{edunov:2018:emnlp_backtrans}
Sergey Edunov, Myle Ott, Michael Auli, and David Grangier. 2018.
\newblock {Understanding Back-Translation at Scale}.
\newblock In \emph{EMNLP}, pages 489--500.

\bibitem[{España-Bonet et~al.(2017)España-Bonet, Ádám Csaba~Varga,
  Barrón-Cedeño, and van Genabith}]{espana2017empirical}
Cristina España-Bonet, Ádám Csaba~Varga, Alberto Barrón-Cedeño, and Josef
  van Genabith. 2017.
\newblock {An Empirical Analysis of NMT-Derived Interlingual Embeddings and
  their Use in Parallel Sentence Identification}.
\newblock \emph{IEEE Journal of Selected Topics in Signal Processing}, pages
  1340--1348.

\bibitem[{Etchegoyhen and Azpeitia(2016)}]{etchegoyhen2016set}
Thierry Etchegoyhen and Andoni Azpeitia. 2016.
\newblock \href {https://doi.org/10.18653/v1/P16-1189} {{Set-Theoretic
  Alignment for Comparable Corpora}}.
\newblock In \emph{ACL}, pages 2009--2018.

\bibitem[{Gehring et~al.(2017)Gehring, Auli, Grangier, Yarats, and
  Dauphin}]{Gehring:2017:fairseq_icml}
Jonas Gehring, Michael Auli, David Grangier, Denis Yarats, and Yann~N Dauphin.
  2017.
\newblock {Convolutional Sequence to Sequence Learning}.
\newblock In \emph{ICML}, pages 1243--1252.

\bibitem[{Gr{\'e}goire and Langlais(2017)}]{gregoire2017bucc}
Francis Gr{\'e}goire and Philippe Langlais. 2017.
\newblock \href {http://aclweb.org/anthology/W17-2509} {{BUCC 2017 Shared Task:
  a First Attempt Toward a Deep Learning Framework for Identifying Parallel
  Sentences in Comparable Corpora}}.
\newblock In \emph{BUCC}, pages 46--50.

\bibitem[{Guo et~al.(2018)Guo, Shen, Yang, Ge, Cer, Abrego, Stevens, Constant,
  Sung, Strope, and Kurzweil}]{guo:2018:wmt_effective}
Mandy Guo, Qinlan Shen, Yinfei Yang, Heming Ge, Daniel Cer, Gustavo~Hernandez
  Abrego, Keith Stevens, Noah Constant, Yun-Hsuan Sung, Brian Strope, and Ray
  Kurzweil. 2018.
\newblock {Effective Parallel Corpus Mining using Bilingual Sentence
  Embeddings}.
\newblock In \emph{WMT}, pages 165--176.

\bibitem[{Hassan et~al.(2018)Hassan, Aue, Chen, Chowdhary, Clark, Federmann,
  Huang, Junczys-Dowmunt, Lewis, Li, Liu, Liu, Luo, Menezes, Qin, Seide, Tan,
  Tian, Wu, Wu, Xia, Zhang, Zhang, and Zhou}]{hassan2018achieving}
Hany Hassan, Anthony Aue, Chang Chen, Vishal Chowdhary, Jonathan Clark,
  Christian Federmann, Xuedong Huang, Marcin Junczys-Dowmunt, William Lewis,
  Mu~Li, Shujie Liu, Tie-Yan Liu, Renqian Luo, Arul Menezes, Tao Qin, Frank
  Seide, Xu~Tan, Fei Tian, Lijun Wu, Shuangzhi Wu, Yingce Xia, Dongdong Zhang,
  Zhirui Zhang, and Ming Zhou. 2018.
\newblock {Achieving Human Parity on Automatic Chinese to English News
  Translation}.
\newblock \emph{arXiv:1803.05567}.

\bibitem[{Khayrallah and Koehn(2018)}]{khayrallah2018impact}
Huda Khayrallah and Philipp Koehn. 2018.
\newblock \href {http://aclweb.org/anthology/W18-2709} {{On the Impact of
  Various Types of Noise on Neural Machine Translation}}.
\newblock In \emph{WNMT}, pages 74--83.

\bibitem[{Koehn(2005)}]{Koehn:2005:mtsummit_eurparl}
Philipp Koehn. 2005.
\newblock Europarl: A parallel corpus for statistical machine translation.
\newblock In \emph{MT summit}.

\bibitem[{Koehn and Knowles(2017)}]{koehn2017six}
Philipp Koehn and Rebecca Knowles. 2017.
\newblock \href {http://www.aclweb.org/anthology/W17-3204} {{Six Challenges for
  Neural Machine Translation}}.
\newblock In \emph{WNMT}, pages 28--39.

\bibitem[{Munteanu and Marcu(2005)}]{munteanu2005improving}
Dragos~Stefan Munteanu and Daniel Marcu. 2005.
\newblock \href {http://www.aclweb.org/anthology/J05-4003} {{Improving Machine
  Translation Performance by Exploiting Non-Parallel Corpora}}.
\newblock \emph{Computational Linguistics}, 31(4):477--504.

\bibitem[{Munteanu and Marcu(2006)}]{munteanu2006extracting}
Dragos~Stefan Munteanu and Daniel Marcu. 2006.
\newblock \href {http://www.aclweb.org/anthology/P06-1011} {{Extracting
  Parallel Sub-Sentential Fragments from Non-Parallel Corpora}}.
\newblock In \emph{ACL}, pages 81--88.

\bibitem[{Ott et~al.(2018)Ott, Edunov, Grangier, and
  Auli}]{Ott:2018:wmt_scale_nmt}
Myle Ott, Sergey Edunov, David Grangier, and Michael Auli. 2018.
\newblock Scaling neural machine translation.
\newblock In \emph{WMT}, pages 1--9.

\bibitem[{Resnik(1999)}]{resnik1999mining}
Philip Resnik. 1999.
\newblock \href {http://www.aclweb.org/anthology/P99-1068} {{Mining the Web for
  Bilingual Text}}.
\newblock In \emph{ACL}.

\bibitem[{Schwenk(2018)}]{schwenk2018filtering}
Holger Schwenk. 2018.
\newblock \href {http://aclweb.org/anthology/P18-2037} {{Filtering and Mining
  Parallel Data in a Joint Multilingual Space}}.
\newblock In \emph{ACL}, pages 228--234.

\bibitem[{Shaw et~al.(2018)Shaw, Uszkoreit, and
  Vaswani}]{shaw:2018:naacl_selfattn}
Peter Shaw, Jakob Uszkoreit, and Ashish Vaswani. 2018.
\newblock {Self-Attention with Relative Position Representations}.
\newblock In \emph{NAACL}, pages 464--468.

\bibitem[{Shi et~al.(2006)Shi, Niu, Zhou, and Gao}]{shi2006dom}
Lei Shi, Cheng Niu, Ming Zhou, and Jianfeng Gao. 2006.
\newblock \href {http://www.aclweb.org/anthology/P06-1062} {{A DOM Tree
  Alignment Model for Mining Parallel Data from the Web}}.
\newblock In \emph{ACL}, pages 489--496.

\bibitem[{Utiyama and Isahara(2003)}]{utiyama2003reliable}
Masao Utiyama and Hitoshi Isahara. 2003.
\newblock \href {http://www.aclweb.org/anthology/P03-1010} {{Reliable Measures
  for Aligning Japanese-English News Articles and Sentences}}.
\newblock In \emph{ACL}.

\bibitem[{Vaswani et~al.(2017)Vaswani, Shazeer, Parmar, Uszkoreit, Jones,
  Gomez, Kaiser, and Polosukhin}]{vaswani2017attention}
Ashish Vaswani, Noam Shazeer, Niki Parmar, Jakob Uszkoreit, Llion Jones,
  Aidan~N Gomez, {\L}ukasz Kaiser, and Illia Polosukhin. 2017.
\newblock {Attention is all you need}.
\newblock In \emph{NIPS}, pages 6000--6010.

\bibitem[{Wu et~al.(2016)Wu, Schuster, Chen, Le, Norouzi, Macherey, Krikun,
  Cao, Gao, Macherey, Klingner, Shah, Johnson, Liu, Łukasz Kaiser, Gouws,
  Kato, Kudo, Kazawa, Stevens, Kurian, Patil, Wang, Young, Smith, Riesa,
  Rudnick, Vinyals, Corrado, Hughes, and Dean}]{wu2016google}
Yonghui Wu, Mike Schuster, Zhifeng Chen, Quoc~V. Le, Mohammad Norouzi, Wolfgang
  Macherey, Maxim Krikun, Yuan Cao, Qin Gao, Klaus Macherey, Jeff Klingner,
  Apurva Shah, Melvin Johnson, Xiaobing Liu, Łukasz Kaiser, Stephan Gouws,
  Yoshikiyo Kato, Taku Kudo, Hideto Kazawa, Keith Stevens, George Kurian,
  Nishant Patil, Wei Wang, Cliff Young, Jason Smith, Jason Riesa, Alex Rudnick,
  Oriol Vinyals, Greg Corrado, Macduff Hughes, and Jeffrey Dean. 2016.
\newblock {Google's Neural Machine Translation System: Bridging the Gap between
  Human and Machine Translation}.
\newblock \emph{arXiv:1609.08144}.

\bibitem[{Xu and Koehn(2017)}]{xu2017zipporah}
Hainan Xu and Philipp Koehn. 2017.
\newblock \href {https://www.aclweb.org/anthology/D17-1319} {{Zipporah: a Fast
  and Scalable Data Cleaning System for Noisy Web-Crawled Parallel Corpora}}.
\newblock In \emph{EMNLP}, pages 2945--2950.

\bibitem[{Ziemski et~al.(2016)Ziemski, Junczys-Dowmunt, and
  Pouliquen}]{ziemski2016united}
Michał Ziemski, Marcin Junczys-Dowmunt, and Bruno Pouliquen. 2016.
\newblock {The United Nations Parallel Corpus v1.0}.
\newblock In \emph{LREC}.

\bibitem[{Zweigenbaum et~al.(2017)Zweigenbaum, Sharoff, and
  Rapp}]{zweigenbaum2017overview}
Pierre Zweigenbaum, Serge Sharoff, and Reinhard Rapp. 2017.
\newblock \href {http://aclweb.org/anthology/W17-2512} {{Overview of the Second
  BUCC Shared Task: Spotting Parallel Sentences in Comparable Corpora}}.
\newblock In \emph{BUCC}, pages 60--67.

\bibitem[{Zweigenbaum et~al.(2018)Zweigenbaum, Sharoff, and
  Rapp}]{zweigenbaum2018overview}
Pierre Zweigenbaum, Serge Sharoff, and Reinhard Rapp. 2018.
\newblock {Overview of the Third BUCC Shared Task: Spotting Parallel Sentences
  in Comparable Corpora}.
\newblock In \emph{BUCC}.

\end{thebibliography}
\bibliographystyle{acl_natbib}

\end{document}